\documentclass[runningheads]{llncs}

\usepackage{eccv}

\usepackage{eccvabbrv}
\usepackage{booktabs}
\usepackage{tabularx}

\usepackage{graphicx}
\usepackage{booktabs}

\usepackage[accsupp]{axessibility}  % Improves PDF readability for those with disabilities.

\usepackage{hyperref}

\usepackage{orcidlink}

\begin{document}

\title{TAU-Agent: An Agentic Retrieval-Augmented Framework for Traffic Anomaly Understanding} 

\titlerunning{TAU-Agent}
\author{
Yuqiang Lin\inst{1}$^*$ \and
Yan Shi\inst{2}$^*$ \and
Sam Lockyer\inst{1} \and
Harish Tayyar Madabushi\inst{1} \and
Adrian Evans\inst{1} \and
Wenbin Li\inst{1} \and
Yinhai Wang\inst{2} \and
Nic Zhang\inst{1}
}

\authorrunning{Y. Lin et al.}

\institute{
University of Bath, Bath BA2 7AY, UK
\and
University of Washington, Seattle, WA 98195, USA\\[1ex]
$^*$Equal contribution
}    

\maketitle

\begin{abstract}

Traffic Anomaly Understanding (TAU) requires models and systems to detect, reason about, and explain anomalous events in transportation videos. To address this challenge, we propose TAU-Agent, an agentic retrieval-augmented framework for traffic anomaly understanding. Given a task query, a central retrieval agent orchestrates two visual perception tools, namely a \textit{Video Captioning Tool} and an \textit{Open-Vocabulary Tracking Tool}, to retrieve and select query-relevant evidence, including captions, temporal intervals, and object trajectories. The selected evidence, together with sampled video frames and the input query, is provided to a supervised fine-tuned vision-language model for final reasoning and answer generation. We evaluate TAU-Agent on both the in-domain and the out-of-domain benchmarks from the AI City Challenge 2026. TAU-Agent achieves scores of 0.6779 on Track 3, 0.3998 on Track 7, and 67.9275 on Track 8, ranking second, twelfth, and fifth, respectively. Code is available at: \url{https://github.com/siri-rouser/TAU-Agent}.


\end{abstract}

\section{Introduction}
\label{sec:intro}
% Logic flow: background in video understanding and traffic anomaly understanding -> problem of current research and we find is the information of anomlay video is not very dense and previous apporch ignore it -> This motivates us to build our system.

%framing the problem: traffic understanding shown interest, through this MLLMs has emerged as the most promising. A MLLM can take information of different formats and process them. This problem has become very promising, a push for accurate automated systems capable of detecting not only clear traffic accidents but all kinds of anomalies. MLLMs are particularly interesting because of their range of application, ideally, MLLM is trained to understand a video and have the ability to make informaed inference, allowing it to be applied to more complex applications. This has lead to MLLM recieving substantial attentention ...

% need a little more justification for necessity; informed descision making for traffic? Use of live monitoring and response systems? Better understanding? Tracking? Expanding? Reduce Cost? Data gathering? - Potential Application should include on-site live traffic anomaly understanding system that allows the control panel acts quick when anomaly happened.(hope that make sense?)

Video understanding has received substantial attention from the computer vision community in recent years. With the rapid development of Multi-modal Large Language Models (MLLMs), e.g.,~\cite{yang2025qwen3, chen2024internvl}, video understanding has evolved beyond conventional perception-oriented tasks, such as video classification, toward comprehensive semantic reasoning. Modern video models are increasingly capable of interpreting complex events, temporal relationships, and interactions in videos~\cite{li2024mvbench, fu2025video}. These advances have substantially expanded the capabilities of video understanding systems, enabling a transition from coarse, high-level perception to fine-grained reasoning over dynamic visual content.

Building upon these advances, Track 3 of the AI City Challenge, Anomalous Events in Transportation, focuses on a specific but challenging subdomain of video understanding: traffic video anomaly understanding. This task requires participants to build a unified system capable of detecting, reasoning about, and explaining anomalous events in transportation videos from multiple perspectives. In practice, the system is expected to answer different types of questions, ranging from binary and multiple-choice questions to temporal grounding and free-text questions. 

Despite the remarkable capabilities of recent video MLLMs, e.g.,~\cite{videollava, clark2026molmo2}, in general video understanding, directly applying them to this challenge remains difficult because the task has several characteristics that are not adequately addressed by conventional video reasoning approaches. Based on our observations of the challenge data, we identify two main challenges. First, the benchmark is highly query-dependent. A single video may contain multiple traffic anomalies as well as numerous normal traffic events, while each question may refer to only one specific anomaly, a particular object or interaction, or even contextual information unrelated to the anomaly itself. Therefore, different questions about the same video may require different temporal segments, objects, and types of evidence. Second, the useful information required for reasoning is inherently sparse in both space and time. Spatially, the target object may occupy only a small region of the frame. Temporally, the event related to the query may occur only briefly within a long video. Therefore, uniform temporal sampling may either miss critical evidence or introduce excessive redundant information that interferes with the reasoning process. Based on these observations, we argue that a more effective problem-solving process should first understand the task-specific query, identify the relevant event both spatially and temporally, and then reason over explicit visual evidence and implicit contextual information to generate a comprehensive answer.

\begin{figure}[t]
  \centering
  \includegraphics[width=\textwidth]{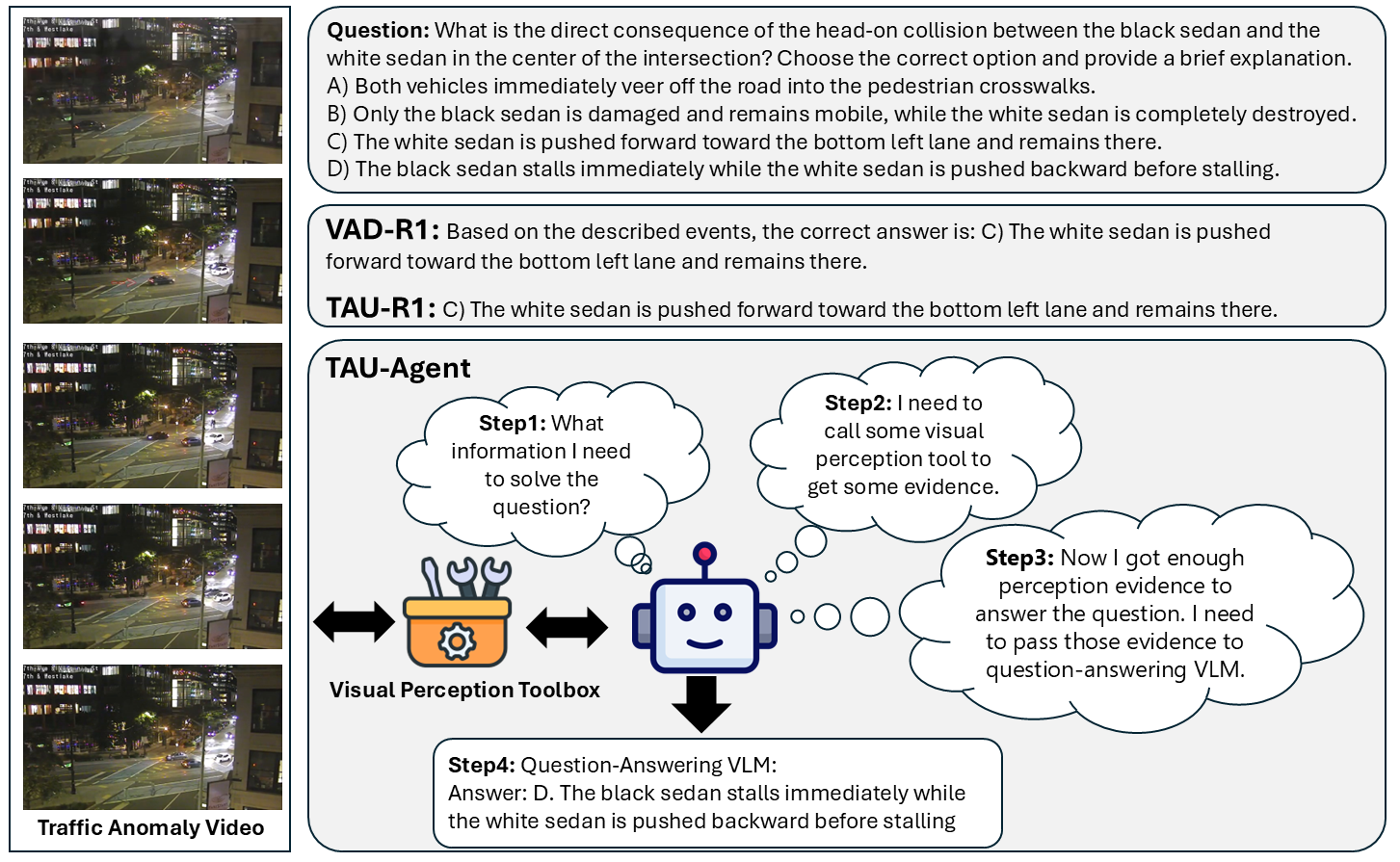}
    \caption{TAU-Agent pipeline example with comparison of other end-to-end unified sampled video anomaly models. TAU-Agent follows multi-stage pipeline to decompose the complex TAU task.}
  \label{fig:example}
\end{figure}

% maybe also mention about the visual processing(captioning + tracking) in the introduction
Recent agentic AI systems~\cite{fan2025videoagent, wang2024videoagent} have demonstrated strong capabilities in decomposing complex reasoning tasks into multiple coordinated stages. We find that the nature of agentic AI systems closely aligns with the problem-solving process described above. Motivated by this observation, we propose \textbf{TAU-Agent}, an agentic framework that addresses the TAU task through the multi-stage process described above, as illustrated in Figure~\ref{fig:example}. Rather than directly reasoning over uniformly sampled video frames, TAU-Agent first interprets the task query to determine the required evidence. It then calls the \textit{video captioning tool} and \textit{open vocabulary tracking tool} to retrieve and select query relevant evidence while also determine query relevant frame range. Finally, the questions, video source and all retrieved evidence are integrated by a fine-tuned question-answering VLM to predict the final answer.

Our contributions are summarized as follows:
\begin{itemize} 
    \item We propose \textbf{TAU-Agent}, an agentic retrieval-augmented framework in which a main agent decomposes complex transportation anomaly understanding tasks and adaptively retrieves query-relevant evidence using two specialized tools: the \textit{Video Captioning Tool} and the \textit{Open-Vocabulary Tracking Tool}. The retrieved visual and textual evidence is then integrated by a supervised fine-tuned VLM to generate the final answer.

   \item We evaluate TAU-Agent on both in-domain and out-of-domain transportation video understanding benchmarks. TAU-Agent ranks second on the in-domain AI City Challenge Track 3 benchmark and achieves competitive performance on the out-of-domain Track 8 PSI-VQA and Track 7 FETV benchmarks, ranking fifth and twelfth, respectively. These results demonstrate the effectiveness of TAU-Agent and provide evidence of its ability to generalize across different traffic-video domains and task formulations.
\end{itemize}
\section{Related Works}
\label{sec:realted_works}
\subsection{Video Anomaly Understanding}
Video anomaly detection traditionally focuses on assigning anomaly scores and localizing unusual temporal segments. Recent vision-language models and MLLMs have broadened this task toward video anomaly understanding (VAU), which additionally involves describing anomalous events and reasoning about their temporal, spatial, and causal context. Such capabilities are particularly important in transportation videos, where anomalies often emerge from evolving interactions among multiple road users.

Existing MLLM-based methods can be broadly organized into three directions. First, language-assisted detection methods use pretrained language or multimodal models to improve anomaly scoring and temporal localization. LAVAD extracts anomaly evidence from frame captions, AnomalyRuler generates scene-specific detection rules, and EventVAD and PrismVAU improve inference through event-level modeling or prompt refinement \cite{zanella2024harnessing,yang2024follow,shao2025eventvad,erreguena2026prismvau}. Second, multitask VAU methods adapt MLLMs with anomaly-oriented instruction data to jointly perform localization, description, and question answering, as explored by VAD-R1, VAD-LLaMA, Holmes-VAD, HAWK, CUVA, Holmes-VAU and TAU-R1 \cite{lv2024vadllama,zhang2024holmesvad,tang2024hawk,du2024cuva,zhang2025holmesvau,huang2025vadr1videoanomalyreasoning,lin2026taur1visuallanguagemodel}. Among these methods, TAU-R1 \cite{lin2026taur1visuallanguagemodel} is, to the best of our knowledge, the only approach that has been specifically evaluated and shown promising results in the transportation domain. Third, reasoning-centric methods move beyond conventional anomaly recognition toward unseen-event generalization and explicit modeling of event structure. LAVIDA targets zero-shot detection of novel anomalies, while VADER reasons over evolving object interactions and causal relations \cite{dai2026lavida,cheng2026vader}. Despite these advances, existing methods often specialize in individual VAU capabilities, leaving temporal localization, spatial grounding, interaction modeling, and causal reasoning insufficiently unified.
\subsection{Agent-based Video Understanding}
Reasoning-intensive video understanding requires models to identify relevant visual evidence, integrate information across time, and perform multi-step inference over events and object interactions. Agent-based methods address this challenge by enabling an LLM or MLLM to iteratively plan, retrieve video segments, invoke perception tools, and refine its prediction. Single-agent systems, including VideoAgent, VideoChat-A1, and DVD, employ iterative shot retrieval, coarse-to-fine temporal search, or caption-based video databases to focus on query-relevant evidence \cite{wang2024videoagent,wang2026videochata1,dvd}. However, their performance remains constrained by incomplete retrieval and the reasoning capacity of a single controller.

Recent work distributes perception and reasoning across multiple agents. VideoMultiAgents combines specialized visual, textual, and graph-based agents, while LVAgent enables multiple MLLMs to retrieve evidence, exchange rationales, and iteratively refine their predictions \cite{kugo2025videomultiagents,chen2025lvagent}. ReAgent-V introduces reward-guided reflection for iterative correction, whereas Symphony decomposes video reasoning into specialized planning, grounding, perception, subtitle-analysis, and reflection roles \cite{zhou2025reagentv,yan2026symphony}. Despite these advances, existing systems often rely on predefined roles and fixed coordination workflows, while errors in evidence retrieval may propagate through subsequent reasoning. Task-adaptive collaboration that jointly improves evidence localization, cross-modal integration, and reasoning reliability therefore remains an open challenge.
\section{Methodology}
\label{sec:methods}
\subsection{Framework Overview}
\begin{figure}[t]
  \centering
  \includegraphics[width=\textwidth]{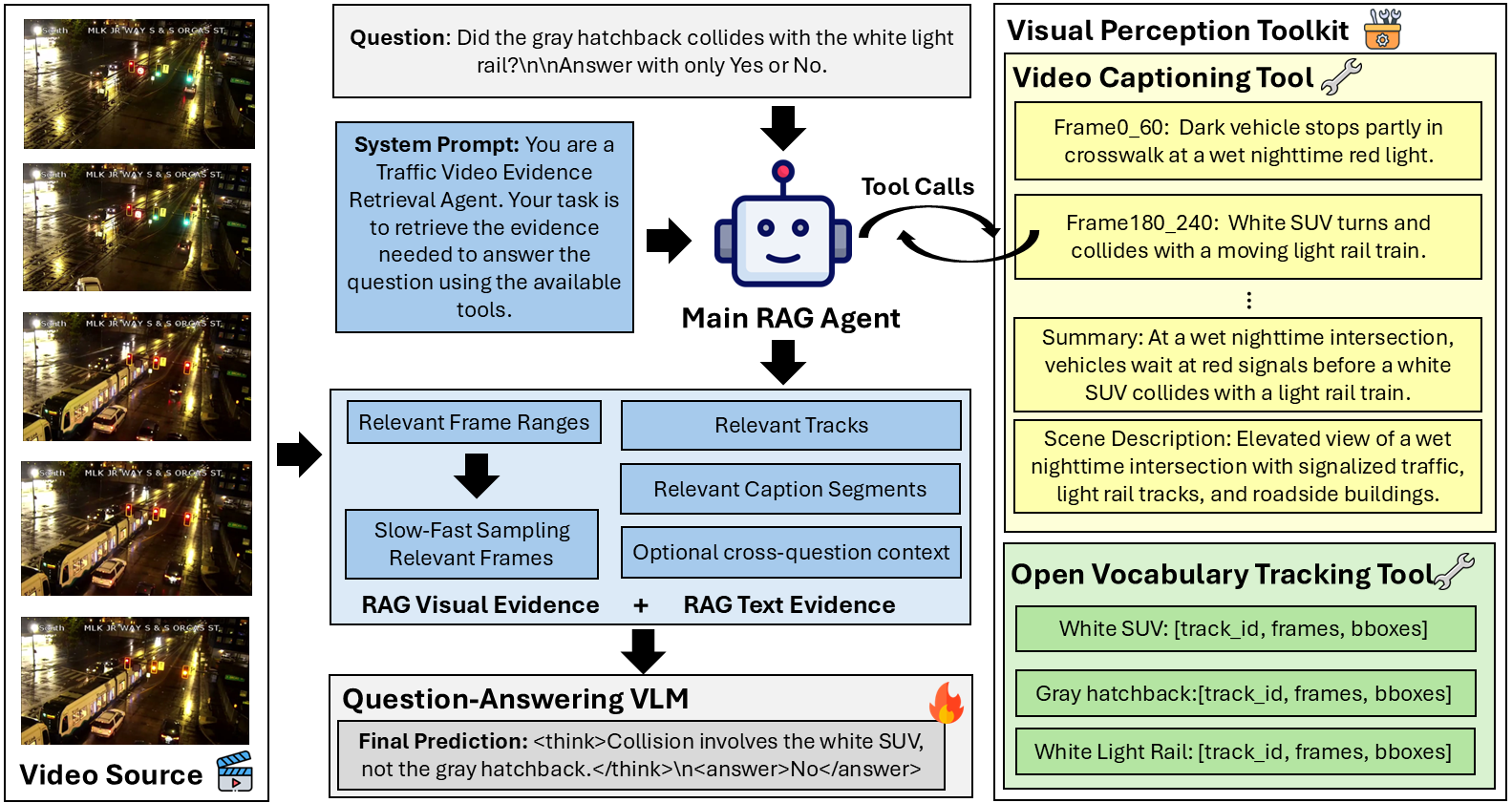}
    \caption{Overview of the TAU-Agent framework. The main agent invokes the \textit{Video Captioning Tool} and the \textit{Open-Vocabulary Tracking Tool} to retrieve and select query-relevant evidence and determine the relevant frame range. The input question, sampled video frames, and retrieved evidence are then passed to a fine-tuned question-answering VLM to generate the final prediction.}
  \label{fig:tau-r1}
\end{figure}

As illustrated in \cref{fig:tau-r1}, the TAU-Agent pipeline begins by providing the input question to the main retrieval-augmented generation (RAG) agent. After interpreting the question, the agent first invokes the \textit{Video Captioning Tool} to obtain a high-level semantic understanding of the video. Based on the question and retrieved captions, the agent conditionally invokes the \textit{Open-Vocabulary Tracking Tool} when object-level evidence is required. The main agent then reasons over the question and all retrieved evidence to select query-relevant captions and object trajectories and determine the relevant frame range. For AI City Challenge Track 3, an optional \textit{Cross-Question Context Agent} further retrieves complementary information from related questions associated with the same video. The selected frame range guides slow-fast sampling of the original video, while the selected captions, object trajectories, and optional cross-question context are incorporated as textual evidence. Finally, the sampled frames, retrieved textual evidence, and input question are jointly passed to a question-answering VLM trained with chain-of-thought (CoT) supervision to generate the final answer.

\subsection{TAU-Agent Framework Design}
\subsubsection{Video Captioning Tool}
Video captions provide a compact semantic representation of long videos, enabling efficient event-level understanding without requiring the main agent to process a large number of visual tokens. Furthermore, high-level semantic descriptions have been shown to facilitate downstream video question answering by providing relevant event representations~\cite{wang2024vamos,zhi2025videoagent2}. Motivated by these observations, we design the \textit{Video Captioning Tool} that extracts textual descriptions at both local and global levels using advanced MLLMs. Specifically, each video is first partitioned into non-overlapping two-second segments. For each segment, frames are uniformly sampled at 2 FPS and provided to MLLMs to generate a segment-level caption describing the local event. The resulting captions are then arranged chronologically and supplied to produce a coherent summary of the entire video, capturing the overall event progression. To further provide scene-level context, four frames are uniformly sampled from the complete video and used to generate a global scene description. Consequently, the Video Captioning Tool produces three different types of textual evidence: (1) temporally localized captions describing fine-grained events, (2) a chronological video summary capturing the overall event evolution, and (3) a global scene description providing holistic contextual information.

\subsubsection{Open Vocabulary Tracking Tool}
\begin{figure}[t]
  \centering
  \includegraphics[width=\textwidth]{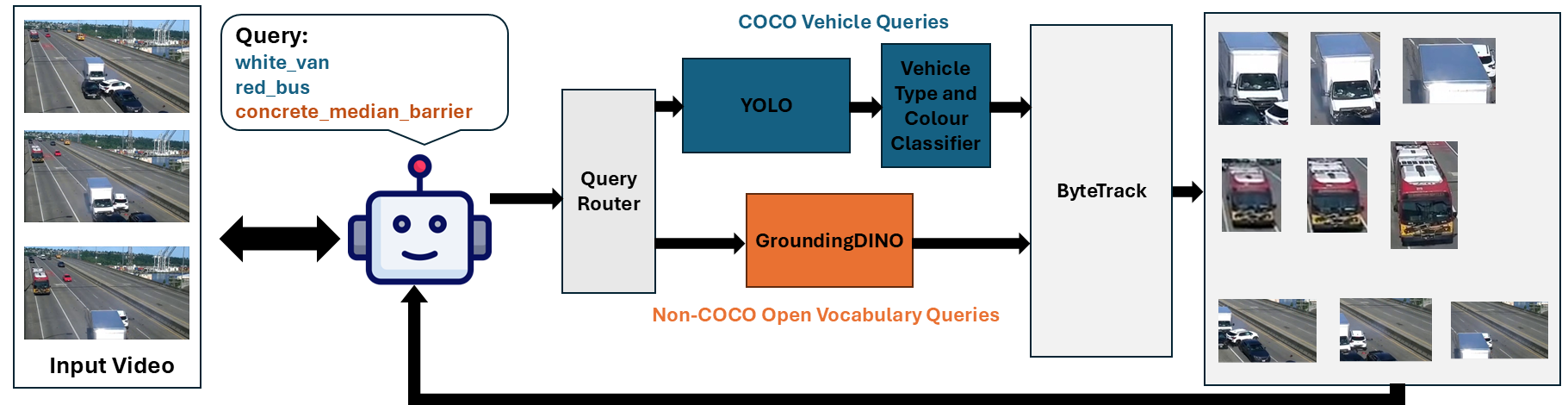}
    \caption{Overview of the hybrid detection and tracking pipeline. Queries shown in blue correspond to traffic-related COCO categories or fine-grained subclasses and are routed through the YOLO-based branch. Queries shown in orange correspond to non-COCO open-vocabulary categories and are processed by GroundingDINO. Detections from both branches are associated across frames using ByteTrack to generate object tracks.}
  \label{fig:object-tracking}
\end{figure}

We consider object trajectories to be a useful source of contextual evidence for video question answering, as they provide fine-grained spatiotemporal information about traffic anomalies and the objects involved. Previous work has also demonstrated that explicit object-centric representations can benefit downstream video question answering~\cite{tang2025can}. Motivated by these observations, we develop an \textit{Open-Vocabulary Tracking Tool} that enables the main agent to extract object-centric information relevant to the input question. To improve detection and tracking robustness, we design the hybrid detection pipeline as illustrated in \cref{fig:object-tracking}. The hybrid pipeline first determines whether the target corresponds to a traffic-related COCO category, a fine-grained vehicle subclass, such as sedan, pickup truck, or SUV, or a non-COCO open-vocabulary category by the given query. Queries associated with COCO vehicle categories and their fine-grained subclasses are routed through a YOLO-based detection branch, where YOLO26~\cite{yolo26} first detects objects from the corresponding coarse category. The detected objects are then processed by vehicle-type and color classifiers to retain instances that satisfy the fine-grained attributes specified in the query. For example, given the query \textit{black SUV}, YOLO first detects all cars, after which only vehicles classified as both black and SUV are retained. In contrast, non-COCO open-vocabulary queries are processed directly by GroundingDINO~\cite{groundingdino}, which produces bounding boxes conditioned on the textual query. Detections from both branches are subsequently passed to ByteTrack~\cite{zhang2022bytetrack}, which associates object instances across frames to generate object tracks. Detection and tracking run in the original FPS and each formatted track is sampled at 1 FPS and capped at 20 observations. Each observation contains the frame index, bounding-box coordinates, object label, and detection confidence. The sampled observations are serialized as textual evidence and passed to the downstream question-answering VLM. This hybrid design provides robust handling of frequently occurring color-and-vehicle-type queries while retaining the flexibility to track objects outside the predefined traffic categories.

\subsubsection{Main Agent Workflow}
\label{sec:workflow}
The main aim of the agent is to retrieve and select evidence relevant to the question query, thereby supporting the downstream video question-answering process. To achieve this goal, we design the lightweight multi-step workflow summarized in \cref{tab:main-agent-workflow}. In this workflow, the agent first interprets the question query and invokes the \textit{Video Captioning Tool} to obtain a high-level semantic understanding of the video and identify potentially relevant temporal segments. Based on the question and the retrieved captions, the agent then determines whether additional object-level evidence is required. If necessary, the agent invokes the \textit{Open-Vocabulary Tracking Tool} with appropriate object queries to retrieve fine-grained trajectory information. The agent subsequently reasons jointly over the question and all retrieved evidence to refine the relevant frame range, select the relevant caption segments and object tracks, and assign a relevance score to each selected item. If the available evidence remains insufficient or ambiguous, the agent can perform additional tool calls before returning the selected evidence to the downstream question-answering VLM.

\begin{table}[t]
    \centering
    \caption{Workflow of the main RAG agent.}
    \label{tab:main-agent-workflow}
    \renewcommand{\arraystretch}{1.15}
    \setlength{\tabcolsep}{5pt}
    \begin{tabularx}{\linewidth}{>{\raggedright\arraybackslash}p{0.42\linewidth} X}
        \toprule
        \textbf{Step and Operation} & \textbf{Description} \\
        \midrule
        \textbf{1. Query Interpretation}
        & Analyze the input question to identify the referenced events, objects, interactions, anomaly, and temporal context. \\

        \textbf{2. Video Caption Retrieval}
        & Invoke the \textit{Video Captioning Tool} to obtain a high-level understanding of the video and retrieve potentially relevant caption segments. \\

        \textbf{3. Temporal Evidence Selection}
        & Use the question and caption evidence to determine candidate frame ranges and select relevant caption segments. \\

        \textbf{4. Object-Track Retrieval}
        & If object-level evidence is required, invoke the \textit{Open-Vocabulary Tracking Tool} to retrieve relevant tracks. \\

        \textbf{5. Evidence Refinement}
        & Jointly reason over the question, captions, and object tracks to refine the selected frame range, select supporting evidence, and assign a relevance score to each selected item. \\

        \textbf{6. Iterative Retrieval}
        & Perform additional tool calls if the retrieved evidence remains insufficient or ambiguous. \\

        \textbf{7. Evidence Output}
        & Return the query-relevant frame range, caption segments, and object tracks to the downstream question-answering VLM. \\
        \bottomrule
    \end{tabularx}
\end{table}

\subsubsection{Cross-Question Context Agent}
In addition to video captions and object tracks, we observe that some questions in AI City Challenge Track 3 are interrelated and may provide complementary contextual information. For example, the question \textit{What is the root cause of the T-bone collision between the white SUV and the black sedan?} provides useful contextual cues for answering another question, such as \textit{Does a T-bone collision occur at the intersection?}. However, such cross-question dependencies are specific to benchmarks in which multiple related questions are associated with the same video and may not generalize to broader transportation anomaly understanding tasks. We therefore implement this capability as an optional \textit{Cross-Question Context Agent} rather than incorporating it into the main RAG agent. Given questions from different tasks associated with the same video, the agent analyzes their wording and extracts three types of contextual evidence: \textit{factual information}, referring to information strongly presupposed by the questions; \textit{potential information}, including weaker hypotheses, candidate events, multiple-choice options, uncertain clues from binary-choice questions, and potentially relevant entities; and \textit{relevant frame ranges}, extracted when questions specify the timestamps of relevant anomalies. When available, the factual and potential information is incorporated into the retrieved textual evidence, while the extracted frame ranges are combined with the frame range selected by the main agent through a union operation. Finally, the resulting cross-question context is combined with the evidence retrieved by the main agent and passed to the downstream question-answering VLM.

\subsubsection{Question-Answering VLM}

The question-answering VLM generates the final answer using the evidence retrieved by the main RAG agent. For visual evidence, we adopt a slow-fast sampling strategy guided by the retrieved frame range. Specifically, the full video is sampled at the default rate to preserve its temporal context, while the query-relevant frame range is sampled at a denser rate to capture fine-grained visual information related to the question. For textual evidence, the five highest-scoring caption segments, the five highest-scoring object tracks, and optional cross-question context are passed to the input question to form an augmented textual prompt. The sampled frames and augmented prompt are then jointly passed to the VLM to generate the final answer.

\subsection{Question-Answering VLM Adaptation}
\subsubsection{Dataset Construction}
We combine the training data provided by AI City Challenge Track 3~\cite{Tang26AICity26} and PSI-VQA~\cite{jing2025psi} as a unified training set. To further improve data diversity and training efficiency, we filter the Track 3 data at the video level by removing highly repetitive or out-of-domain videos. Specifically, we manually identify and remove 1,843 normal videos from So-TAD~\cite{chen2025so}, 228 normal videos from the HTV dataset~\cite{chan2005probabilistic}, and 128 normal videos from \texttt{barbados\_challenge}~\cite{barbados_traffic_analysis_2025}. These videos contain highly repetitive scenes captured by the same cameras and therefore provide limited additional visual diversity. We also remove 99 videos from the ShanghaiTech dataset~\cite{liu2018ano_pred}, included as part of VAD-R1~\cite{huang2025vadr1videoanomalyreasoning}, because their content is unrelated to traffic anomalies. The remaining videos are processed offline by TAU-Agent to retrieve query-relevant evidence. Compared with the standard TAU-Agent workflow described in \cref{sec:workflow}, we introduce an additional evidence-validation step during training-set construction to reduce noise introduced by the retrieval and evidence-selection process. Specifically, the ground-truth answer and its corresponding CoT reasoning trace are provided to the RAG agent as training-time context. The agent verifies whether the selected captions, object tracks, and frame ranges support the target answer and reasoning process. If the selected evidence is insufficient or inconsistent with the target reasoning, the agent is instructed to revise its selection. The ground-truth answer and CoT trace are used only for evidence validation and are not included in the evidence provided to the question-answering VLM. The validated captions and object tracks are then combined with the cross-question context to form the augmented textual evidence. Finally, the resulting query-specific evidence is stored locally and loaded directly during training, avoiding repeated tool calls and improving training efficiency.

\subsubsection{Task-Specific Prompt Engineering}
The ten tasks in AI City Challenge Track 3 have different reasoning objectives and output requirements. Most tasks require the model to detect, localize, reason about, or explain specific anomalous events, whereas \textit{scene description} and \textit{video summarization} have distinct objectives. Specifically, \textit{scene description} requires a detailed and objective description of the static traffic environment, while \textit{video summarization} requires a chronological account of the main events and their development. We therefore organize the tasks into three groups: anomaly-focused question answering, scene description, and video summarization. A separate system prompt is designed for each group to align the VLM with the corresponding reasoning objective and response format. In addition to using task-specific system prompts, we adapt the retrieved evidence to the requirements of each task group. For anomaly-focused question answering and video summarization, the model receives the query-specific evidence selected by the RAG system. For scene description, event-specific captions and object trajectories may introduce irrelevant details or overemphasize individual events. Therefore, we provide only the global scene description generated by the \textit{Video Captioning Tool} as auxiliary textual evidence. The sampled video frames remain available to the VLM for all task groups.

\subsubsection{VLM Training Strategy}
We adopt parameter-efficient supervised fine-tuning using LoRA~\cite{hu2022lora} to adapt a pretrained VLM to the unified TAU task. During training, the question, sampled video frames, retrieved textual evidence, and task-specific system prompt are jointly provided to the VLM. The model is supervised using the corresponding CoT reasoning trace and final answer. This enables the model to learn task-specific reasoning patterns across diverse traffic anomaly scenarios while aligning its responses with the target answer formats.
\section{Experiments}
\label{sec:experiment}

We evaluate TAU-Agent on the in-domain TAR benchmark~\cite{Tang26AICity26} and two out-of-domain benchmarks: the FishEye Traffic Violation (FETV) dataset~\cite{Tang26AICity26} and PSI-VQA~\cite{jing2025psi}. We compare the performance of TAU-Agent with that of other teams participating in the AI City Challenge.

\subsection{Implementation Details}

The \textit{Video Captioning Tool} uses models from the Gemini family~\cite{team2023gemini}. Specifically, \texttt{gemini-3.5-flash} is used for the training data, providing a balance between cost and quality. The main agent and the optional cross-question context agent use \texttt{gpt-5.4-2026-03-05}, selected due to its strong reasoning ability.

For the question-answering VLM, we adopt Qwen3-VL-8B~\cite{yang2025qwen3} as the base model and perform parameter-efficient supervised fine-tuning using LoRA. We set the LoRA rank $r$ to 128, the scaling factor $\alpha$ to 256, and the dropout rate to 0.03. The maximum number of input frames is set to 100. The full video is sampled at 2 FPS, while the query-relevant frame range is sampled more densely at 4 FPS. We train the model for two epochs using a learning rate of $5 \times 10^{-5}$ with effective batch size of $8$. All training and evaluation experiments are conducted using two NVIDIA RTX PRO 6000 Blackwell GPUs.

\subsection{Tar Test}
\subsubsection{Overview}

TAR Test is the official in-domain benchmark of AI City Challenge Track 3. It contains 80 traffic-surveillance videos and covers ten tasks: event verification, event verification with explanation, multiple-choice question answering, multiple-choice question answering with explanation, open-ended question answering, scene description, video summarization, temporal localization, causal linkage, and event description. Binary-choice and multiple-choice questions are evaluated using accuracy, while the remaining open-ended tasks are evaluated using BERTScore F1. Temporal localization is evaluated using mean Intersection over Union (mIoU); however, this task is excluded from the final overall evaluation by the AI City Challenge committee.

\subsubsection{Dataset Pre-Processing}

We preprocess TAR Test using the same RAG pipeline employed to construct the training evidence. For the \textit{Video Captioning Tool}, we replace \texttt{gemini-3.5-flash} with \texttt{gemini-3.1-pro-preview} to obtain more accurate and detailed captions for the test videos. For each question, the original video, retrieved evidence, and task-specific prompt are jointly provided to the fine-tuned question-answering VLM to generate the initial prediction.

\subsubsection{Result Post-Processing}

To further improve performance on TAR Test, we apply three benchmark-specific, context-aware post-processing strategies to refine the initial predictions: (1) \textit{Context-Aware Binary Answer Refinement}, (2) \textit{Context-Aware Multiple-Choice Alignment}, and (3) \textit{Context-Aware Free-Text Consensus Reranking}. These strategies are motivated by the observation that questions associated with the same video are often interrelated, allowing predictions for one question to provide useful context for verifying or refining another. For \textit{Context-Aware Binary Answer Refinement}, we generate five candidate responses for each BCQ and BCQ-Open question. One candidate is generated through greedy decoding with a temperature of 0, while the remaining four are sampled with a temperature of 0.7. Majority voting is then applied to obtain the initial binary prediction. We empirically observe that paired BCQ and BCQ-Open questions typically contain one ``yes'' answer and one ``no'' answer. When the voted predictions do not follow this pattern, the VLM first reconsiders each question independently using predictions from other questions associated with the same video as additional contextual evidence. If the inconsistency remains, the paired questions are jointly provided to the VLM, which is instructed to assign one ``yes'' answer and one ``no'' answer. For \textit{Context-Aware Multiple-Choice Alignment}, we use the same candidate-generation and majority-voting configuration. Since each MCQ and MCQ-Open pair asks an equivalent question but presents the answer options in a different order, we map both predictions to their corresponding option content and evaluate their consistency. If the predicted answers differ, the VLM reconsiders each question using predictions from other questions associated with the same video as contextual evidence. If the inconsistency remains, the MCQ-Open prediction is aligned with the option content selected for the corresponding MCQ. Finally, \textit{Context-Aware Free-Text Consensus Reranking} is applied to the remaining open-ended tasks. For each question, we generate five candidate responses using predictions from related questions about the same video as contextual evidence. One candidate is generated through greedy decoding with a temperature of 0, while the remaining four are sampled with a temperature of 0.7. The final response is selected through medoid reranking based on pairwise BERTScore F1 similarity. Specifically, the candidate with the highest average similarity to all other candidates is selected as the consensus answer.
\subsubsection{Main Results}
We compare our TAU-Agent with other top-ranked submissions on the in-domain TAR Test benchmark as shown in \cref{tab:tar_leaderboard}. Overall, our submission ranks second with a mean score of 0.6779, only 0.0009 below the top-ranked entry. Additionally, our method also achieves the highest scores among the listed submissions on causal linkage, temporal description, and video summarization, while matching the best results on BCQ and MCQ. 

\begin{table}[t]
\centering
\caption{Results on the in-domain TAR leaderboard. Our submission is shown in \textit{italics} and best item are bold.}
\label{tab:tar_leaderboard}
\resizebox{\linewidth}{!}{
\begin{tabular}{c l c c c c c c c c c c}
\toprule
Rank & Team & Mean & BCQ & MCQ & BCQ OE & MCQ OE &
Open QA & Causal & Scene & Temporal & Summary \\
\midrule
1 & 25
& 0.6788 & 1.0000 & 0.9500 & 0.6686 & 0.9693
& 0.4986 & 0.5310 & 0.4373 & 0.5137 & 0.5409 \\

\textit{2} &
\textbf{Ours} &
\textit{0.6779} &
\textit{\textbf{1.0000}} &
\textit{\textbf{0.9500}} &
\textit{0.6685} &
\textit{0.9176} &
\textit{0.5150} &
\textit{\textbf{0.5503}} &
\textit{0.4314} &
\textit{\textbf{0.5164}} &
\textit{\textbf{0.5516}} \\

3 & 309
& 0.6748 & 0.9750 & 0.9500 & 0.6660 & 0.9520
& 0.5215 & 0.5253 & 0.4445 & 0.4975 & 0.5416 \\

4 & 270
& 0.6741 & 0.9750 & 0.9500 & 0.6663 & 0.9530
& 0.5177 & 0.5181 & 0.4439 & 0.4980 & 0.5453 \\

5 & 60
& 0.6703 & 0.9750 & 0.9500 & 0.6663 & 0.9530
& 0.5177 & 0.5181 & 0.4095 & 0.4980 & 0.5453 \\
\bottomrule
\end{tabular}}
\end{table}

\subsection{FETV}
\subsubsection{Overview}
FETV is the official out-of-domain benchmark of AI City Challenge Track 7. It contains 200 short video clips extracted from the Fisheye8K~\cite{Gochoo_2023_CVPR} source videos and presents two major forms of domain shift: (1) out-of-domain visual perception and (2) out-of-domain task formulation. In terms of visual perception, all FETV videos are captured using fisheye cameras, which introduce substantial geometric distortion compared with conventional traffic-surveillance videos. In terms of task formulation, rather than evaluating multiple video question-answering tasks, FETV requires the model to predict 12 structured attributes and generate a free-form caption from the source video. Those structured attributes include date, time, violation type, violator type, color, initial position, final position, initial lane, final lane, intersection type, weather, and lighting condition, with several attributes selected from predefined candidate values. Different metrics are used to evaluate these outputs. Categorical attributes are evaluated using macro-averaged F1, the date field is evaluated by exact matching, and the time field is considered correct if the prediction falls within seven seconds of the ground-truth timestamp. The free-form caption is evaluated using normalized CIDEr and BERTScore. The final FETV score combines normalized CIDEr, BERTScore, and MacroF1 with weights of 0.25, 0.25, and 0.50, respectively.

\subsubsection{Dataset Pre-Processing}
To adapt our framework to the task format required by FETV, we consolidate the requirements of all 12 target attributes into a unified question and instruct the model to produce a single JSON-formatted response that can be directly parsed for evaluation. Using this constructed question as the query, we apply the standard TAU-Agent workflow described in \cref{sec:workflow} to retrieve relevant captions and object tracks. To improve object-detection robustness under fisheye distortion, we fine-tune YOLO on the Fisheye8K dataset and incorporate the resulting model into the \textit{Open-Vocabulary Tracking Tool}. For the \textit{Video Captioning Tool}, we use \texttt{gemini-3.1-pro-preview} to generate more accurate video captions. Finally, the constructed question, original video, and retrieved evidence are jointly passed to the same fine-tuned question-answering VLM used for the other benchmarks to generate the final prediction.

\subsubsection{Results}
We compare TAU-Agent with other top-ranked submissions on the out-of-domain FETV test set in \cref{tab:fetv_leaderboard}. TAU-Agent ranks 12th with an overall score of 0.3998, comprising a description score of 0.3513 and a categorical mean score of 0.4484. These results demonstrate that the unified TAU-Agent framework can be transferred to a substantially different visual domain and output format. However, a performance gap remains between our submission and the highest-ranked methods, particularly in structured attribute prediction. One possible reason is that our question-answering VLM is trained primarily on conventional traffic videos and video question-answering tasks, with limited task-specific adaptation to fisheye imagery and structured JSON prediction. Further adaptation to the FETV domain and individual target attributes may improve performance.

\begin{table}[t]
\centering
\caption{Results on the FETV leaderboard. Our submission is shown in \textit{italics}.}
\label{tab:fetv_leaderboard}
\resizebox{0.75\linewidth}{!}{
\begin{tabular}{c l c c c}
\toprule
Rank & Team & Final Score & Description & Categorical Mean \\
\midrule
1 & 30
& 0.4891 & 0.4171 & 0.5612 \\

2 & 219
& 0.4889 & 0.4166 & 0.5612 \\

3 & 139
& 0.4884 & 0.4411 & 0.5358 \\

\textit{12} &
\textbf{Ours} &
\textit{0.3998} &
\textit{0.3513} &
\textit{0.4484} \\

13 & 61
& 0.3997 & 0.3476 & 0.4518 \\
\bottomrule
\end{tabular}
}
\end{table}

\subsection{PSI\_VQA}

\subsubsection{Overview}
PSI-VQA is an optional out-of-domain benchmark in AI City Challenge Track 8. It contains 40 egocentric dashcam videos from the PSI 2.0 dataset~\cite{jing2025psi}, focusing on pedestrian-crossing scenarios. Compared with the in-domain CCTV data, PSI-VQA introduces two major domain shifts: from overhead surveillance views to egocentric dashcam views, and from traffic anomaly understanding to pedestrian-intent reasoning. The benchmark includes four tasks aligned with those in TAR Test: binary classification of pedestrian-crossing intent, open-ended articulation of ambiguous-intent cues, multiple-choice identification of relevant cues, and temporal localization of driver-decision-critical intervals. These tasks are evaluated using Macro-F1, cue-level F1, accuracy, and mean temporal Intersection over Union (mIoU), respectively. The normalized task scores are equally weighted to obtain the final overall score.
\subsubsection{Dataset Pre-Processing}

Similar to the construction of the training evidence, we preprocess PSI-VQA using the standard TAU-Agent workflow described in \cref{sec:workflow}. The cross-question context is reconstructed deterministically from the released questions without any additional API calls. For each question, the original video, retrieved evidence, and task-specific prompt are jointly provided to the same fine-tuned question-answering VLM used for the other benchmarks to generate the initial prediction.

\subsubsection{Result Post-Processing}
Unlike TAR Test, where the cross-question context is used for all tasks, we find it to be a double-edged signal on PSI-VQA and therefore apply it \emph{task-selectively}. For the open-ended cue-articulation task (Open QA), the context enumerates the candidate crossing-intent cues that the reference answer is drawn from, so retaining it substantially improves cue-F1. For the binary (BCQ) task, however, the same context is harmful: the corresponding videos carry no multiple-choice options, so the context reduces to a one-sided ``the pedestrian may intend to cross'' restatement that biases the prediction toward a single label. It likewise nudges the multiple-choice (MCQ) prediction toward the option surfaced first in the context. We therefore retain the cross-question context only for Open QA and withhold it for BCQ and MCQ, keeping only the visual evidence. No further post-processing is applied to the PSI-VQA predictions.

\subsubsection{Main Results}

Table~\ref{tab:psi_results} reports the results on the out-of-domain PSI-VQA benchmark. TAU-Agent achieves an overall score of 67.9275 and ranks fifth on the leaderboard. Notably, TAU-Agent obtains an Open QA Cue-F1 score of 0.7791, the highest among the listed submissions and 0.1117 higher than the second-best result. This result indicates that the framework performs particularly well in identifying and articulating visual cues related to ambiguous pedestrian-crossing intentions.
\begin{table}[t]
\centering
\caption{Results on the out-of-domain PSI-VQA benchmark. Our submission is shown in \textit{italics} and best item are bold.}
\label{tab:psi_results}
\resizebox{\linewidth}{!}{
\begin{tabular}{c l c c c c c c}
\toprule
Rank & Team & Final & BCQ F1 & BCQ Acc. & Open QA F1 & MCQ Acc. & Temp. mIoU \\
\midrule
1 & 220
& 70.8947 & 0.7084 & 0.7273 & 0.5833 & 0.7912 & 0.7529 \\

2 & 30
& 70.6397 & 0.7084 & 0.7273 & 0.5833 & 0.7912 & 0.7427 \\

3 & 257
& 69.0698 & 0.6136 & 0.6182 & 0.6674 & 0.7912 & 0.6906 \\

4 & 84
& 68.2765 & 0.7136 & 0.7273 & 0.6206 & 0.6813 & 0.7155 \\

\textit{5} &
\textbf{Ours} &
\textit{67.9275} &
\textit{0.6167} &
\textit{0.7091} &
\textit{\textbf{0.7791}} &
\textit{0.7253} &
\textit{0.5960} \\

\bottomrule
\end{tabular}
}
\end{table}

\section{Conclusion}
\label{sec:conclusion}
In this work, we introduced TAU-Agent, an agentic retrieval-augmented framework that coordinates visual perception tools to retrieve query-relevant evidence for traffic anomaly understanding. TAU-Agent ranked second on the in-domain AI City Challenge Track 3 benchmark, twelfth on the out-of-domain Track 7 benchmark, and fifth on the out-of-domain Track 8 benchmark. These results demonstrate in-domain performance and provide evidence that the framework can generalize across different traffic-video domains and task formulations. Future work will extend TAU-Agent to streaming and real-time video understanding, enabling more efficient deployment in practical transportation scenarios.

\section*{Acknowledgements}
Yuqiang Lin and Sam Lockyer are supported by a scholarship from the EPSRC Centre for Doctoral Training in Advanced Automotive Propulsion Systems (AAPS) under project EP/S023364/1.

\bibliographystyle{splncs04}
\bibliography{main}
\end{document}